\def\year{2020}\relax
\documentclass[letterpaper]{article} 
\usepackage{aaai20}  
\usepackage{times}  
\usepackage{helvet} 
\usepackage{courier}  
\usepackage[hyphens]{url}  
\usepackage{graphicx} 
\usepackage{CJKutf8}
\usepackage{subfig}
\usepackage{booktabs}
\usepackage{amsmath}
\usepackage{comment}
\usepackage{hyperref}
\usepackage{algorithm}
\usepackage{algorithmic}
\usepackage{latexsym}
\usepackage{microtype}
\usepackage{hypcap} 
\usepackage{graphicx}  
\newcommand{\citet}[1]{\citeauthor{#1} \shortcite{#1}}
\title{Diversity by Phonetics and its Application in Neural Machine Translation}
\author{Abdul Rafae Khan and Jia Xu}
 
\begin{document}

\maketitle

\begin{abstract}
We introduce a powerful approach for Neural Machine Translation (NMT), whereby, during training and testing, together with the input we provide its phonetic encoding and the variants of such an encoding.  This way we obtain very significant improvements up to 4 BLEU points over the state-of-the-art large-scale system. The phonetic encoding is the first part of our contribution, with a second being a theory that aims to understand the reason for this improvement. Our hypothesis states that the phonetic encoding helps NMT because it encodes a procedure to emphasize the difference between semantically diverse sentences. We conduct an empirical geometric validation of our hypothesis in support of which we obtain overwhelming evidence. Subsequently, as our third contribution and based on our theory, we develop artificial mechanisms that leverage during learning the hypothesized (and verified) effect phonetics. We achieve significant and consistent improvements overall language pairs and datasets: French-English, German-English, and Chinese-English in medium task IWSLT'17 and French-English in large task WMT'18 Bio, with up to 4 BLEU points over the state-of-the-art. Moreover, our approaches are more robust than baselines when evaluated on unknown out-of-domain test sets with up to a 5 BLEU point increase.

\end{abstract}

\section{Introduction}

This work presents a novel framework for Neural
Machine Translation (NMT) using phonetic information
`computed' by human interaction throughout
the evolution of spoken language. Our overarching goal is to improve machine translation accuracy. 
We view social interaction as a computational
device that generates precomputed knowledge. 
We systematically study the possible reasons for getting improved machine translation systems and based on this we construct artificial analogs of the phonetic encoding and apply these ideas to a wide range of human language pairs and domains. Our empirical study shows an overwhelming improvement over the state-of-the-art. In what follows we discuss our motivation, give an overview of
our methods, and summarize our results. 

Machine translation is one of the areas that we have seen remarkable advancement with the revival of the neural networks. Deep neural networks are believed to automatically learn the underlying features. To that end, neural networks typically have an extensive number of parameters to be estimated from the data. The data can have a biased domain or noise such as typos. It is a challenging task to rely purely on neural networks to extract all hidden features in NMT. Besides, we are not aware of a well-known general mathematical explanation about the effectiveness of the shortcomings/bottlenecks of artificial neural networks. Adding another representation as input potentially allows a more straightforward network structure. Therefore, we ask:

``Can we find a new representation of natural language data beyond the text that is less sensitive to styles or errors (without an additional input source)?''

\noindent
\emph{Phonetics is another language representation (besides text). } Let us consider how humans communicate a message. Our first recourse, far earlier than any written language, was to encode our thoughts in sound~\cite{blevins2004evolutionary}. 
Similar phenomena happen for individuals in the first language acquisition when children start with talking than reading.
Phonology, during human language development, carries semantic meanings, see~\cite{tyler1996interaction,beaver2007semantics}. These studies coincide with neural discoveries about the correlation between phonology and semantics in the human brain ~\cite{wang2016neural,Amenta2017}. The phonological structure is realized through phonetic implementation as in~\cite{cohn2014interface}. 

\noindent
\emph{What is phonetics?}  A phonetic algorithm (encoding) is an algorithm for indexing words by their pronunciation.  Table~\ref{tab-phoneticsexamples} shows examples of phonetics: Pinyin and Soundex. We view these encodings as many-to-one functions that map multiple words to one (formally speaking it is a projection of the text). We introduce Soundex, NYSIIS, Metaphone for Western languages, and use Pinyin for Chinese. In addition to phonetics, we consider other forms of `projections' for Chinese and in particular a Logogram encoding and Wubi.

\noindent
\emph{Can the coupled representation of phonetics and text benefit MT?} 
We use speech in the form of a phonetic sequence as an independent language representation of text. Note that this can be efficiently computed by the given text input -- i.e.~throughout our paper we do not use any additional input/information. Furthermore, we do not need any additional labeled dataset or other sources. Phonetics has a smaller vocabulary than written text. Adding a second representation of the foreign sentences after their text form will increase the tolerance to noise and other lexical variations.

\begin{figure}[t]
    \centering
    \hspace{-1cm}\vspace{-1cm}
    \includegraphics[width=0.9\textwidth]{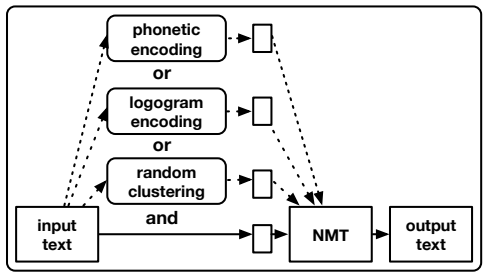}
        \vspace{-5cm}
    \caption{\small{Workflow: how to apply our approaches into NMT. No new data needed (such as labeling or input source). BPE and embedding are in the empty box.}}
    \label{fig:workflow}
\end{figure}

\noindent
\emph{How to use phonetics in MT? } 
We give this new form of phonetic sentence representation together with its written text form as an input when training and decoding the neural networks. We use (1) concatenation of phonetic and word sequences; (2) multi-source encoding of phonetics and text form in words.
Figure~\ref{fig:workflow} shows the workflow of how we apply our methods. We first apply phonetic encoding, logogram, or random clustering to the foreign sentences.  Then we apply Byte-Pair-Encoding and learn a word embedding (marked as empty boxes). Source and target embedding are trained jointly. Finally, we concatenate or combine with multi-source encoder the embedding of the original text and the phonetics to feed into NMT. We treat the NMT as a black-box and thus it is technically easy to adapt to any new NMT or preprocessing tools.

\begin{table}[t]
\begin{center}\scalebox{.7}{
\begin{tabular}{l|l|l}
function & output &input\\\hline
Pinyin &xiao4&  \begin{CJK*}{UTF8}{gbsn}笑(smile), 校(school), 孝(filial), 效(imitate)\end{CJK*} \\
&shi4& \begin{CJK*}{UTF8}{gbsn}氏(surname),事(matter),市(city),视(vision) \end{CJK*} \\
Soundex & B300& body,but,bad\\
&S120& speak,space,suppose,speech\\
&C600& car,care,chair,cherry,choir,cry,crow,core\\
\end{tabular}}\caption{\small{Phonetics is a many-to-one  function.}}\label{tab-phoneticsexamples}
\end{center}
\end{table}

We achieved substantial and consistent increases over the state-of-the-art on all language pairs with which we experimented. 
We conducted extensive experiments and achieved up to \textbf{4 BLEU points} on the medium scale IWSLT'17 and the large scale WMT'18 Biomedical task over the state-of-the-art in translation directions of English-German, German-English, English-French, French-English, and Chinese-English. In particular, this led to a higher accuracy even on an arbitrary test set whose distribution is unknown at training. We verified that our approaches are more robust on French-English experiments with about a \textbf{5 BLEU point} improvement on a foreign test set from an unknown domain. Our approach is general and can potentially benefit any language with phonetic encodings and \textit{any} NMT system.

\emph{Why phonetics help significantly in NMT?} We investigate the structure of the phonetic encoding aiming to explain the translation improvements. We performed a systematic empirical analysis to understand the effect of phonetic encodings. We introduce and verify with three quantitative analyses 
our hypothesis of \textit{semantic diversity by phonetics} stated as: 

\begin{quote}
\textit{``One phonetic representation usually corresponds 
to characters/words that are semantically diverse.''}
\end{quote}

\emph{From phonetics to artificial encodings based on our hypothesis.} 
To explain why the phonetics diversity hypothesis in NMT we develop an artificial encoding. This is a new random clustering algorithm that casts words or characters into classes randomly. 
Note that a random mapping is still a mapping (i.e.~one word is mapped to a random class, but the same word seen later in the dataset will be mapped to the same class). Random clustering groups more semantically diverse words/characters than typical clusterings such as K-means. This procedure is similar to how words with different meanings map to one phonetic representation according to our hypothesis. The distribution on the random cluster size follows the number of words in each Metaphone. We uniform sample words randomly for each cluster. 
Our experimental results show that the random clustering achieves comparable improvements to phonetic encodings. This finding aligns with the empirical justification of our hypothesis on why phonetic encoding improves NMT. 

This work contains two areas of study: phonetic encoding and random clustering, where the former inspires the latter per our hypothesis. Here are our main contributions:

\begin{enumerate}
\item \textit {Phonetic and logogram encodings. }
We introduce phonetic and logogram encoding to enrich the text representation of languages. We experimented with English, French, German, and Chinese on various encodings such as  Soundex, NYSIIS, Metaphone, Pinyin, and Wubi. These encodings, when used as auxiliary inputs, improve state-of-the-art NMT with up to 4 and 5 BLEU points on an in-domain and an unknown out-of-domain test set, respectively. 

\item \textit {The phonetics diversity hypothesis. }
We empirically analyze our hypothesis. We find and verify that a phonetic representation corresponds to words with diverse semantic meanings. Phonetics is a function that groups semantically different words. Note that this finding could be of independent interest to the field of Linguistics.

\item \textit {Random clustering. }
Inspired by our hypothesis, we introduce random clustering. The considerable improvements over state-of-the-art NMT systems can be explained by the validity of our hypothesis and how this affects the training of a much more accurate but simpler Neural Networks.
\end{enumerate}

In the remainder of the paper, we first introduce \emph{phonetic and logogram encodings}. Then, we study why adding them (each derived from the original text) improves NMT and propose our hypothesis. Subsequently, we introduce an artificial method, \emph{random clustering} to generalize  \emph{text encoding}. Finally, we empirically demonstrate that all these approaches significantly boost the NMT accuracy and robustness.

\section{Background}
\label{sec:background}

NMT is an approach to MT using neural networks, which takes as an input a source sentence $(x_1,..,x_t,..,x_{I})$ and generates its translation $(y_1,..,y_{t'},..,y_{I'})$, where $x_t$ and $y_{t'}$ are source and target words respectively.
NMT models with attention have three components, namely, an encoder, a decoder, and an attention mechanism.
The encoder summarizes the meaning of the input sequence by encoding it with a bidirectional recurrent neural network (RNN). 
We apply the sequence-to-sequence learning architecture by~\cite{Gehring17}, where the encoder and decoder states are calculated using convolutional neural networks (CNNs). 

\section{Phonetic Encodings}

A phonetic algorithm is used to index words by their pronunciation. We apply the phonetic algorithm to each word in a sentence and output a sequence of phonetic encodings. 

\subsection{Soundex}

Soundex is a widely known phonetic algorithm for indexing names by sound and avoids misspelling and alternative spelling problems. It maps homophones to the same representation despite minor differences in spelling~\cite{soundex:07}.  Continental European family names share the 26 letters (A to Z) in English. 
Soundex clusters the letter with exceptions.
For example, the Soundex key letter code clusters `b, f, p, v' to `1', and `c, g, j, k, q, s, x, z' to `2', and `d, t' to `3'.

\subsection{NYSIIS}

The New York State Identification and Intelligence System Phonetic Code (NYSIIS)  is a phonetic algorithm devised in 1970~\cite{rajkovic2007adaptation}.  It features an accuracy increase of 2.7\% over Soundex and takes special care to handle phonemes that occur in European and Hispanic surnames by adding rules to Soundex. For example, if the last letters of the name are `EE' then these letters are changed to `Yb'.

\subsection{Metaphone}

Metaphone~\cite{philips1990hanging} is another algorithm that improves on earlier systems such as Soundex and NYSIIS. The Metaphone algorithm is significantly more complicated than the others because it includes special rules for handling spelling inconsistencies and for looking at combinations of consonants in addition to some vowels.

\subsection{Hanyu Pinyin}

Hanyu Pinyin (Pinyin) is the official romanization system for Standard Chinese in mainland China. Pinyin, which means `spelled sound', was developed to teach Mandarin. One Pinyin corresponds to multiple Chinese characters. One Chinese word is usually composed of one, two, or three Chinese characters.

\subsection{Logogram Encoding: Chinese Wubi}

The Wubizingxing (Wubi or Wubi Xing) is a Chinese character input method primarily used to efficiently input Chinese text with a keyboard. 
The Wubi method decomposes a character based on its structure rather than its pronunciation. 
Every character can be written with at most 4 keystrokes including -, $\mid$, \begin{CJK*}{UTF8}{gbsn}丿\end{CJK*}, hook, and \begin{CJK*}{UTF8}{gbsn}丶\end{CJK*} with various combinations. 

\section{Random Clustering}\label{subsec-randomclustering}

\begin{algorithm}[tb]
\caption{\small{Random Clustering.}}\label{alg-rancluster}
 \algsetup{linenosize=\tiny}
  \scriptsize
\label{alg:algorithm}
\textbf{Input}: translation units\\
\textbf{Parameter}: baseline encoding\\
\textbf{Output}: mapping of units to clusters\\
\begin{algorithmic}[1] 
\STATE perform a phonetic or logogram encoding as baseline
\FOR {each unique code in the baseline encoding vocabulary }
\STATE $Z=$ ``how many units are mapped''
\STATE uniformly random sample $Z$ units to form a new cluster
\ENDFOR
\STATE \textbf{return} 
\end{algorithmic}
\end{algorithm}

Driven by our hypothesis, which we will be elaborate in Section~\ref{sec-hyps}, we further introduce an artificial method to encode text that simulates 'natural' encoding (i.e. phonetics and logogram).
We call this random clustering as described in Algorithm~\ref{alg-rancluster}.
We cluster words (or characters) uniformly at random.
The cluster size follows the distribution of how many words/characters are associated with each phonetic algorithm, here Metaphone. For example, in Chinese, each Pinyin is a cluster. Each cluster size is the same as the number of characters mapped to a Pinyin, and the number of clusters equals the number of unique Pinyins. 

\section{Hypothesis}\label{sec-hyps}

\paragraph{Hypothesis:} \emph{One phonetic representation (for example, Pinyin in Chinese) usually corresponds to characters/words that are semantically diverse.}

At first, this hypothesis may seem counter-intuitive. However, it is made to reduce ambiguity in oral communication, because, otherwise, humans would not be able to communicate effectively due to confusion. For example, red (Pinyin: `hong') and green (Pinyin: `lv') in Chinese appear in similar contexts.
It seems plausible to think that part of the development of phonetics is that one re-uses the same sound when context can be used to distinguish among multiple interpretations. For example, 
`to' versus `two.'

How do we set up experiments to verify this?

We test our hypothesis using geometric interpretations of semantics, precisely, word embeddings~\cite{bengio2014word}. 
Intuitively, an embedding~\cite{mikolov2013distributed,arora2016linear}  preserves pairwise semantic distances. For instance, the two words/characters are close if they are semantically similar and far away otherwise.
For instance, see the work of ~\citet{zouzias2010low,molitor2017remarks} about volume preserving embeddings, which formalizes the concept of this term. For example, if we have a set of words, and all the words correspond to a Pinyin, then the points themselves may mean nothing, but the distances among the points are our focus. Typically in geometry, three points in space are sufficient to quantify a volume. 
We embed each word or characters from Chinese-English translation data (in Section~\ref{sec-exp}) into 100 dimensions and then project this embedding into two dimensions using PCA.
Algorithm~\ref{alg-convexhull2} describes how we compute a smooth convex hull of points. The convex hull of a Pinyin is the convex hull of embeddings of all words or characters pronounced with this Pinyin. 

\begin{algorithm}[t]
 \algsetup{linenosize=\tiny}
  \scriptsize
\caption{\small{$c$: Smoothed Convex Hull of Points.}}
\label{alg:algorithm}
\textbf{Input}: points (embedded $R^2$ vectors) in a cluster \\
\textbf{Parameters}: $\beta$: threshold; $r$: radius\\
\textbf{Output}: The convex hull's vertices \\
\begin{algorithmic}[1] 
\FOR{for each point}
\STATE{draw a circle with $r$}
\IF{the total number of points in the circle is less than $\beta$}
\STATE remove this point
\ENDIF
\ENDFOR
\STATE \textbf{return} the convex hull 
\end{algorithmic}\label{alg-convexhull2}
\end{algorithm}

\begin{figure*}[t]
    \centering
    \subfloat[Plot 1-4: words (black dots) of Soundex `B631', `E455', `V536', `O550'respectively.]{{\includegraphics[width=7.4cm]{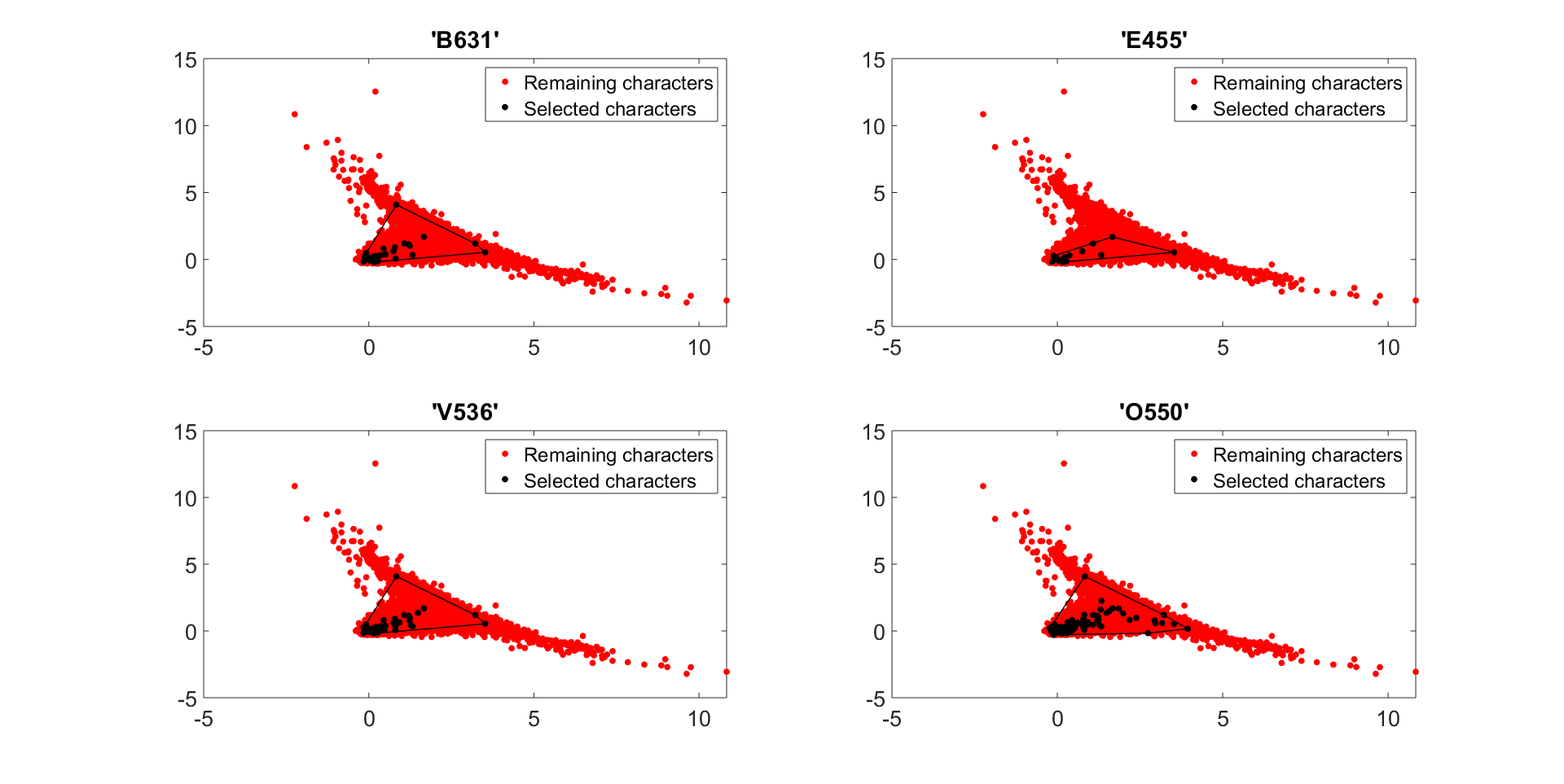} }}%
    \qquad
    \subfloat[ Plot 1-4: all Chinese characters (black dots) of Pinyin `gen4', `si4', `guo2', `ju4'respectively.]{{\includegraphics[width=7.4cm]{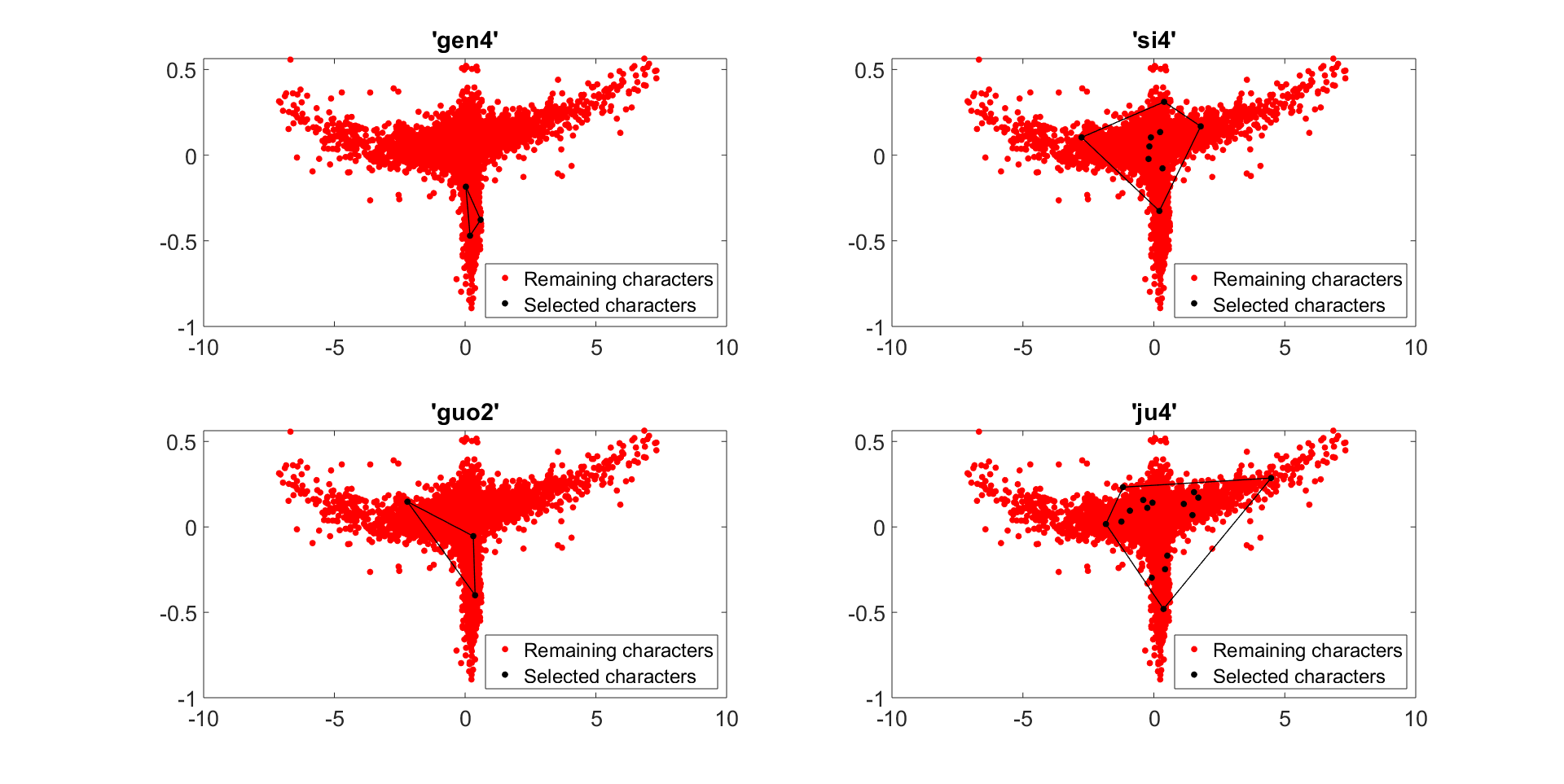} }}%
    \caption{\small{Same pronounced words/Chinese characters  have distributed meaning in semantic  space (red dots).}}%
   \label{fig:convexhull_all}
\end{figure*}

\begin{figure*}[t]
    \centering
    \subfloat[K-Means, Soundex, and random clustering coverage speed by adding words (black dots) of group: The convex hull volume (black lines) of Soundex and random clustering cover the space (red dots) faster than K-Means by increasing the number of groups .]{{\includegraphics[width=7.4cm]{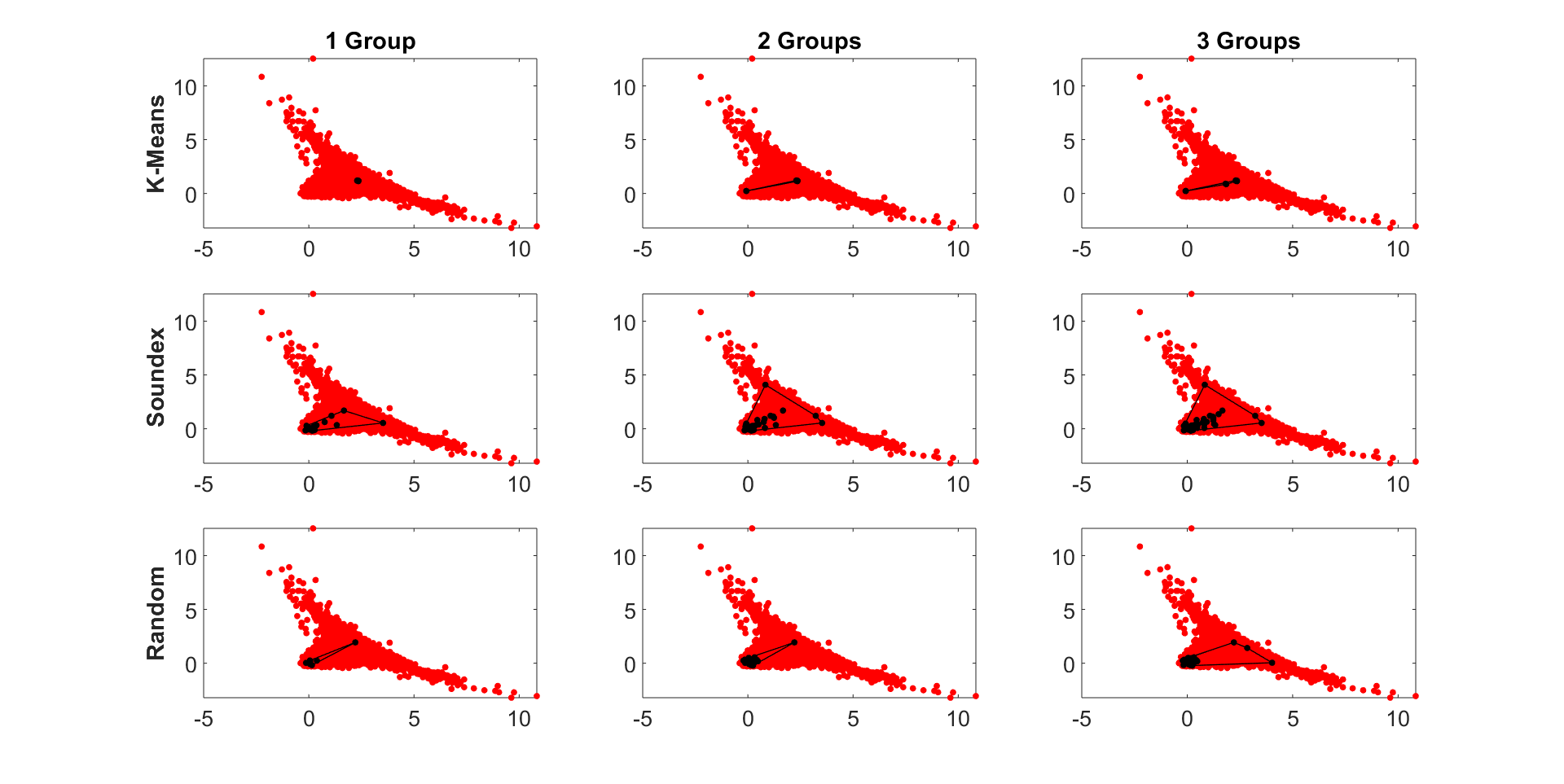} }}%
    \qquad
    \subfloat[K-Means, Pinyin, and random clustering coverage speed by adding characters (black dots) of each cluster or Pinyin: The convex hull volume (black lines) of Pinyin and random clustering cover the space (red dots) faster than K-Means by increasing the number of groups.]{{\includegraphics[width=7.4cm]{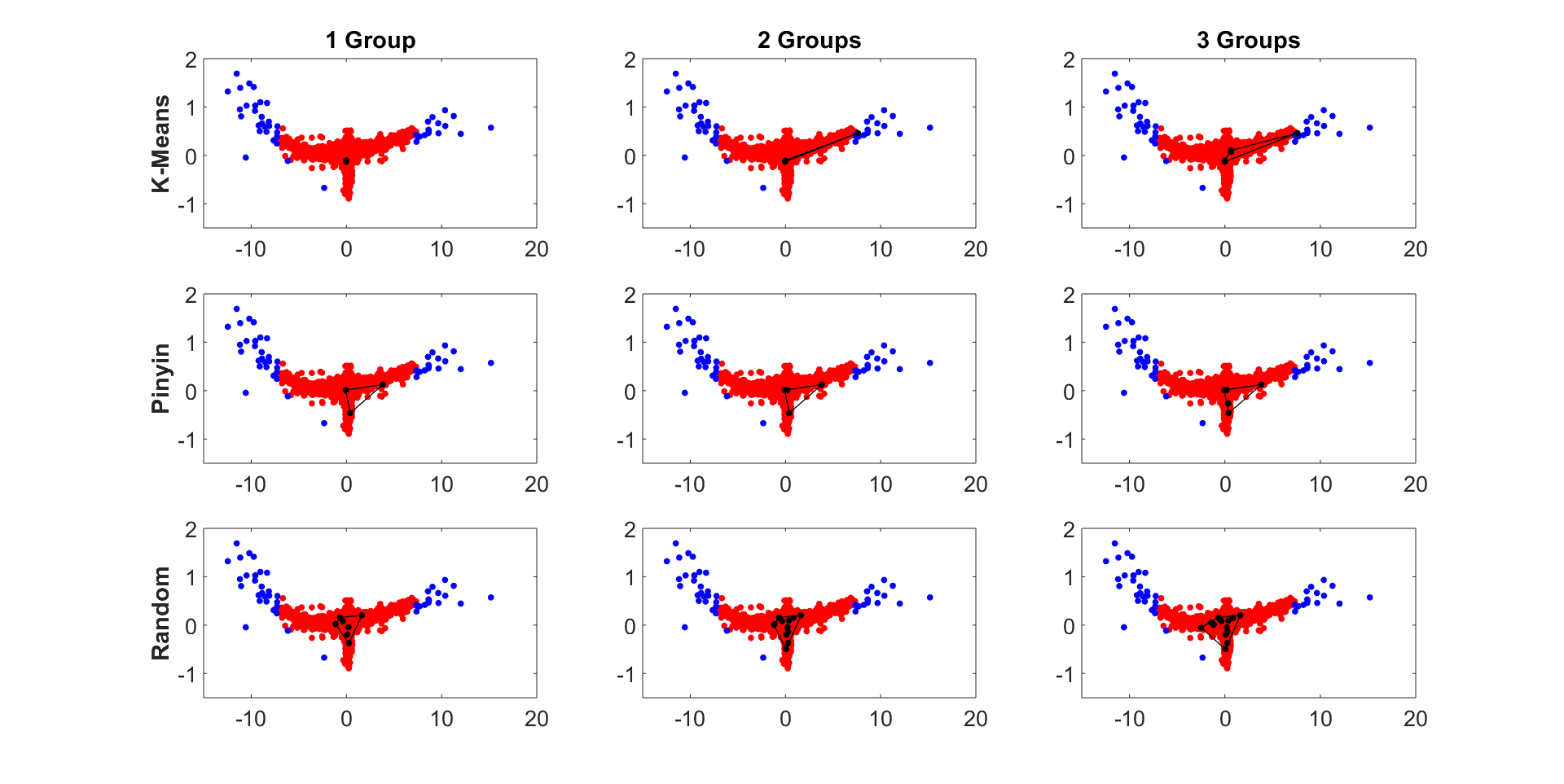} }}%
    \caption{\small{Coverage speed when adding groups one by one.}}%
   \label{fig:convexhull_iterations}
\end{figure*}

\paragraph{Observations.} Figure~\ref{fig:convexhull_all} shows all embedded Chinese characters in red dots, and black dots are the Chinese character(s) of one random Pinyin in each plot. We can see that characters with the same pronunciation tend to have distributed meaning - that is, well-distributed over the Euclidean plane.

In Figure~\ref{fig:convexhull_iterations}, we measure the convex hull (the smallest convex set that contains all points - implemented in Matlab) of all characters. We exclude the outliers (blue dots) by removing all points that are encircled along with less than $\beta$ other points in a ball of radius $r$.  The first plot shows the hull enclosing characters of one random group (either cluster or Pinyin). The second plot shows the addition of characters from a second random group to the first group, analogously for the second plot. The convex hull volume (here, 2D volume) increases as we add groups. 
We can see that Soundex, Pinyin, and random grouping covers the space faster than the K-Means clustering when we increase the number of groups. 

\begin{figure}[t]
    \centering
    \includegraphics[width=0.5\textwidth]{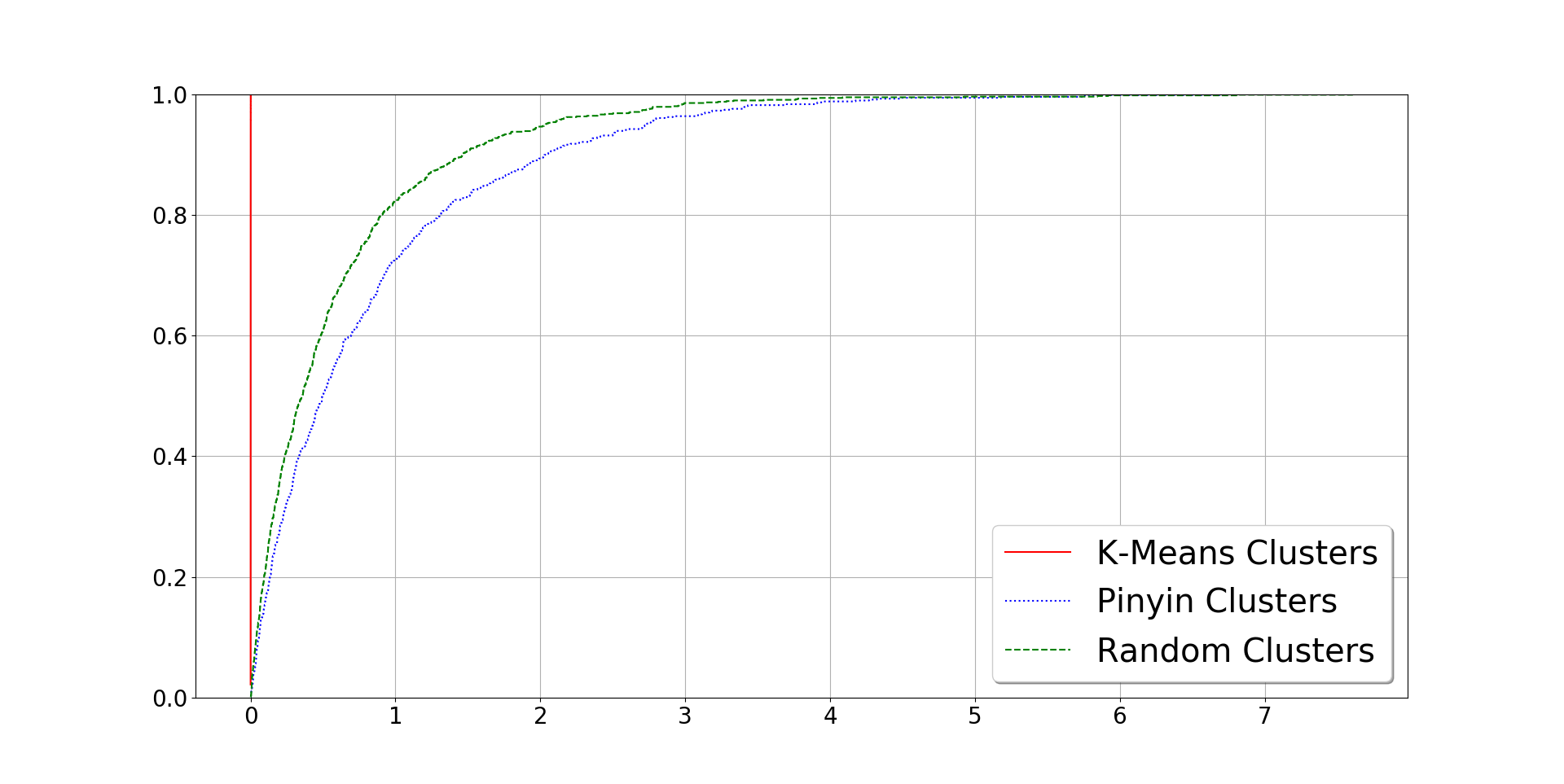}
    \caption{\small{CDF of the convex hull volume of characters in each group (cluster or Pinyin) using three methods. K-Means has a very small convex hull volume in each group. The volume of Pinyin, and random clustering are close, but Pinyin is even larger.}}
    \label{fig:histogram}
\end{figure}

\paragraph{Quantitative verifications.} These are carried out with three experiments.  
First, Figure~\ref{fig:histogram} shows the empirical \textbf{CDF} of the convex hull volume of characters of each Pinyin, random clustering, and K-Means clustering, where the x-axes indicate the volume, and y-axes indicate the frequency. Random clustering and Pinyin grouping have a larger volume than K-Means, respectively. For each group, Pinyin is slightly better distributed (more widespread) than uniformly random clustering, and both of these are better distributed than K-Means. This phenomenon is quite exciting and is probably due to the isoperimetry of the uniform random sampling for these data points.

Second, we define the \textbf{concentration factor}  as 
\begin{eqnarray}
\Gamma({\mathbf{p}_1^K}_{1}^{I_k}) = \frac{\sum_{k=1}^K \lVert \mathbf{C}_k-\frac{\sum_{k=1}^K{\mathbf{C}_k}}{K}\lVert_2 }{ \sum_{k=1}^K \sum_{i=1}^{I_k}  \lVert \mathbf{p}_{ki}-\mathbf{C}_k \rVert_2} \nonumber,
\end{eqnarray} 
where $\mathbf{C}_k=\frac{\sum_{i=1}^{I_k} \mathbf{p}_{ki}}{I_k}$.  $\mathbf{p}_{ki}$ is the $i-$th point in group $k$ (either cluster or Pinyin). The smaller the value, the more distributed the points are in each cluster. The concentration factor $\Gamma$ is 9350 for K-Means, 3.783 for Pinyin, 1.476 for the random clustering in Chinese; 3543K for K-Means, 0.3674 for Soundex, and 0.0191 for random clustering.

\begin{center}
\begin{algorithm}[t]
 \algsetup{linenosize=\tiny}
  \scriptsize
   \caption{Density Measure.}
   \label{alg:density}
\begin{algorithmic}
   \STATE {\bfseries Input:} Set of points $P$, clusters $Q$, nearest neighbor $i$
   \STATE {\bfseries Output:} Density $density$
   \STATE $cluster=$ 5 random clusters from $Q$
   \STATE $s=$ set of wordvectors of all words in $cluster$
   \STATE Remove outliers from $P$ using $\beta=0.3$ \& $r=10$
   \STATE $C$ = corner points of the convexhull of $P$
   \FOR{$i=1$ {\bfseries to} $m-1$}
        \STATE $q=$ $|C|$ random numbers between $0$ \& $1$
        \FOR{$i=1$ {\bfseries to} $|C|$}
            \STATE $C^{\prime}_i = C_i*\frac{q_i}{\sum_{k=1}^{|C|}q_k}$
        \ENDFOR
        \STATE $hullpt = sum(C^{\prime})$
        \STATE $density = 0$
         \REPEAT
            \STATE $density +=$ KNN$(hullpt,chosen,i)$ \COMMENT{KNN outputs distance}
        \UNTIL{$noChange$ is $true$}
   \ENDFOR
\end{algorithmic}
\end{algorithm}
\end{center}

Finally, we define the \textbf{density measure} as in Algorithm~\ref{alg:density}, which intuitively seems to be a more robust test. For each point $x$ in the smoothed convex hull of all words/characters, let $D_i(x)$ be the distance between $x$ and the $i$-th nearest neighbor of $x$ in the space $X$. We then look at either the maximum of $D_i(x)$ over all $x$, or the average. Choosing a larger $i$ captures the `density' of the point-set at larger scales, which is a parameter that can be tuned to be more robust against noise. We numerically integrate the convex hull surface by randomly sampling the points, which are a  linear combination of the convex hull corner weighted uniformly at random. Table~\ref{tab-density} shows the density results, which are consistent with the CDF in Figure~\ref{fig:histogram}. Pinyin is the most well-distributed, then random clustering, followed by K-Means. 

\begin{table}[tb]
\begin{center}\scalebox{.9}{
\begin{tabular}{c|c|c|c|c|c|c}
& \multicolumn{3}{c}{Max}&\multicolumn{3}{c}{Sum}\\
Method&1&2&3&1&2&3\\\hline
K-Means & 0.26 & 0.29 & 0.35 & 19.3 & 26.2 & 31.6 \\
Random & 0.18 & 0.21 & 0.22 & 15.5 & 19.3 & 21.2 \\
Pinyin & 0.08 & 0.21 & 0.22 & 7.19 & 19.4 & 20.9 \\
\end{tabular}}
\caption{\small{Density result (Converge threshold: 0.001; 1, 2, 3 nearest neighbour). Smaller values, more distributed words in each cluster.}}\label{tab-density}
\vspace{-0.5cm}
\end{center}
\end{table}

\section{Experiments}\label{sec-exp}

\paragraph{Adding auxiliary information:} First, we convert all the words or characters in the source sentences (in the training, development, and test set) into phonetic or other encodings. 
Then, we segment those sequences of encodings with the Byte Pair Encoding compression algorithm~\cite{Sennrich16,gage1994new} (BPE). We learn a new embedding on this encoded training data only (for example, Metaphone encodings). Next, the embedded vectors are either concatenated or multi-source encoded with the original sentence embedding (after BPE) or used alone as inputs to the encoder of the CNN neural translator. 

\subsection{Datasets and Vocabularies} 

We carried out experiments for five translation directions:  Chinese to English (ZH-EN), English to French (EN-FR), French to English (FR-EN), English to German (EN-DE), and German to English (DE-EN). We used the IWSLT'17~\cite{iwslt17} and the WMT'18 Biomedical training data. 
Figure~\ref{tab-voc1} shows vocabulary statistics on source/target tokenized text~\cite{chinesetokenizer} before and after applying  encodings~\cite{jellyfishpackage}. 
We apply a BPE with 89K and 16K~\cite{denkowski2017stronger} operations for FR, 89K for DE, and 18K operations for ZH.

\begin{table}[t]

\centering
\scalebox{.7}{
\begin{tabular}{c|cccc||c|c}
Source& EN&EN&FR&DE&Source&ZH\\
Target&FR&DE&EN&EN&Target&EN\\\hline
Source(Words) & 54k & 51k & 73k & 119k &Source(Words)&94k\\
Target & 73k & 119k & 54k & 51k &Target&54k\\\hline
Soundex & 10k & 10k &  - & 16k&Pinyin&1k\\
NYSIIS & 38k & 36k &  43k & 99k&Wubi&4k\\
Metaphone &  36k & 34k &  37k & 94k &&\\\hline
W+Soundex &  58k & 55k &  - & 124k&W+Pinyin&95k\\
W+NYSIIS &  84k & 80k &  108k & 206k&W+Wubi&97k\\
W+Metaphone & 83k & 79k &  104k & 203k&&\\
\end{tabular}}
\caption{\small{Vocabulary sizes before/after encodings.}}\label{tab-voc1}
\end{table}

\subsection{Translation Results} \label{sec-nmtresult}

For each encoding scheme, we carried out two experiments, one with only the encoded sentences and another one with the \textit{source sentence concatenated with the encoded sentence}. For example, W+Soundex means the source sentence in words concatenated with all words converted into Soundex as the input to NMT. As the Soundex algorithm does not support French, we do not have its results for French. We evaluate translation quality with BLEU implemented by~\citet{multibleu}.

\begin{table*}[t]
\begin{center}\scalebox{0.9}{
\begin{tabular}{c|ccccccc}
Coding&FR-EN$_{(89k)}$&FR-EN$_{(16k)}$& EN-FR$_{(89k)}$&EN-FR$_{(16k)}$&DE-EN$_{(89k)}$&EN-DE$_{(89k)}$\\\hline
Baseline: Words~\cite{Gehring17} (W)&35.01&36.21& 34.37&36.78&27.79&25.12\\\hline
Soundex &-&-&27.44&27.41&20.89&21.19\\
NYSIIS &30.87&31.22&31.36&31.06&25.76&18.90\\
Metaphone &29.83&30.43&31.10&30.77&23.61&21.92\\\hline
W+Soundex &-&-&35.88&36.80&27.54&24.97\\
W+NYSIIS &\textbf{35.44}&\textbf{37.33}&35.10&37.23&28.40&25.37\\
W+Metaphone &35.09&37.04&\textbf{36.08}&\textbf{37.95}&\textbf{28.99}&25.00\\
W+random clustering &35.02&36.84&35.47 &37.07&28.21&\textbf{25.58}\\\hline
Baseline: Words~\cite{Vaswani17} ($\omega$) &36.81&&&&&\\\cline{1-2}
$\omega$$\odot$Metaphone &\textbf{37.30}&&&&&
\end{tabular}}\caption{\small{Translation results in BLEU[\%] on a medium task IWSLT. Dev: combined test'13, 14, 15; Test: test'17. BPE operations: ${89k}$, ${16k}$. +: concatenation; $\odot$: multi-source encoding.}}\label{tab-result}
\end{center}
\end{table*}

Table~\ref{tab-result} shows that encoding as an auxiliary input (concatenated with the original sentence) significantly improves the translation quality in all language directions in our experiments.  W+Metaphone indicates adding Metaphone to the word-based NMT baseline, which gives the best results for EN-FR and DE-EN, with an improvement of 1.71 and 1.2 in BLEU points, respectively. 
In our experiments, random clustering consistently improves over baselines on all languages. The non-uniform random clustering method in algorithm~\ref{alg-rancluster}  achieves a higher BLEU score of 37.95\% than a uniform random clustering after tuning on the cluster size. For EN-FR ($16k$ BPE operations) data in Table 5, we uniform sample words randomly for each cluster. We get the BLEU score of 37.74\%, 37.77\%, 37.38\%, and 37.63\% when setting the number of clusters to be 20\%, 40\%, 60\%, and 80\% of the vocabulary size (63615 words), i.e. the average cluster size to be 5, 2.5, 1.6, 1.25, respectively. 

However, for most languages, the best codings are phonetic ones. 
Since linguistic information is typically language-dependent, some phonetic encoding may be more suitable for certain languages than others. NYSIIS handles phonemes that occur in European and Hispanic surnames. Thus, it performs best in French. Metaphone is an advanced algorithm with spelling variations and inconsistencies. Hence, it works best for  English and German (both Germanic languages).

Table~\ref{tab-cnresult} shows the results of the ZH-EN  translation system (BPE $18k$ operations).
We apply Pinyin~\cite{pinyinwubipackage}, and Pinyin segmented into letters, and Wubi encoding~\cite{pinyinwubipackage}. We achieve significant improvement over the baseline by adding auxiliary information: 0.87 BLEU points with Pinyin, 1.68 BLEU points with Pinyin in letters, and 1.11 BLEU points with Wubi, respectively. Randomly clustering on Chinese characters and words both improve the baseline with 1.49 and 1.47 BLEU points, respectively, a more significant improvement than that of the K-Means clustering. 

Additionally, we experimented on English-French and French-English with a large dataset: the WMT'18 Biomedical task that contains more than 2 million sentences. We show the translation results in Table \ref{tab:exp_bio}. We achieved a significant improvement of 2.21 BLEU points for English-French and 4.27 BLEU points for French-English, respectively.

\begin{table}[t]
\begin{center}\scalebox{0.9}{
\begin{tabular}{c|cc}
Coding&EN-FR&FR-EN\\\hline
Baseline: Words ~\cite{Gehring17}&31.10 & 28.19 \\\hline
W+Soundex &32.96 &-\\
W+NYSIIS &32.80 & \textbf{32.46 (+4.27)}\\
W+Metaphone &\textbf{33.31} & 31.95\\
\end{tabular}}\caption{\small{Translation results in BLEU[\%] on large task WMT'18 Bio; Dev: Khresmoi; Test: EDP'17 test. BPE: ${89k}$.} }\label{tab:exp_bio}
\end{center}
\end{table}

\subsection{Model Complexity} \label{sec-complexity}

\begin{figure}
    \centering
    \includegraphics[width=0.5\textwidth]{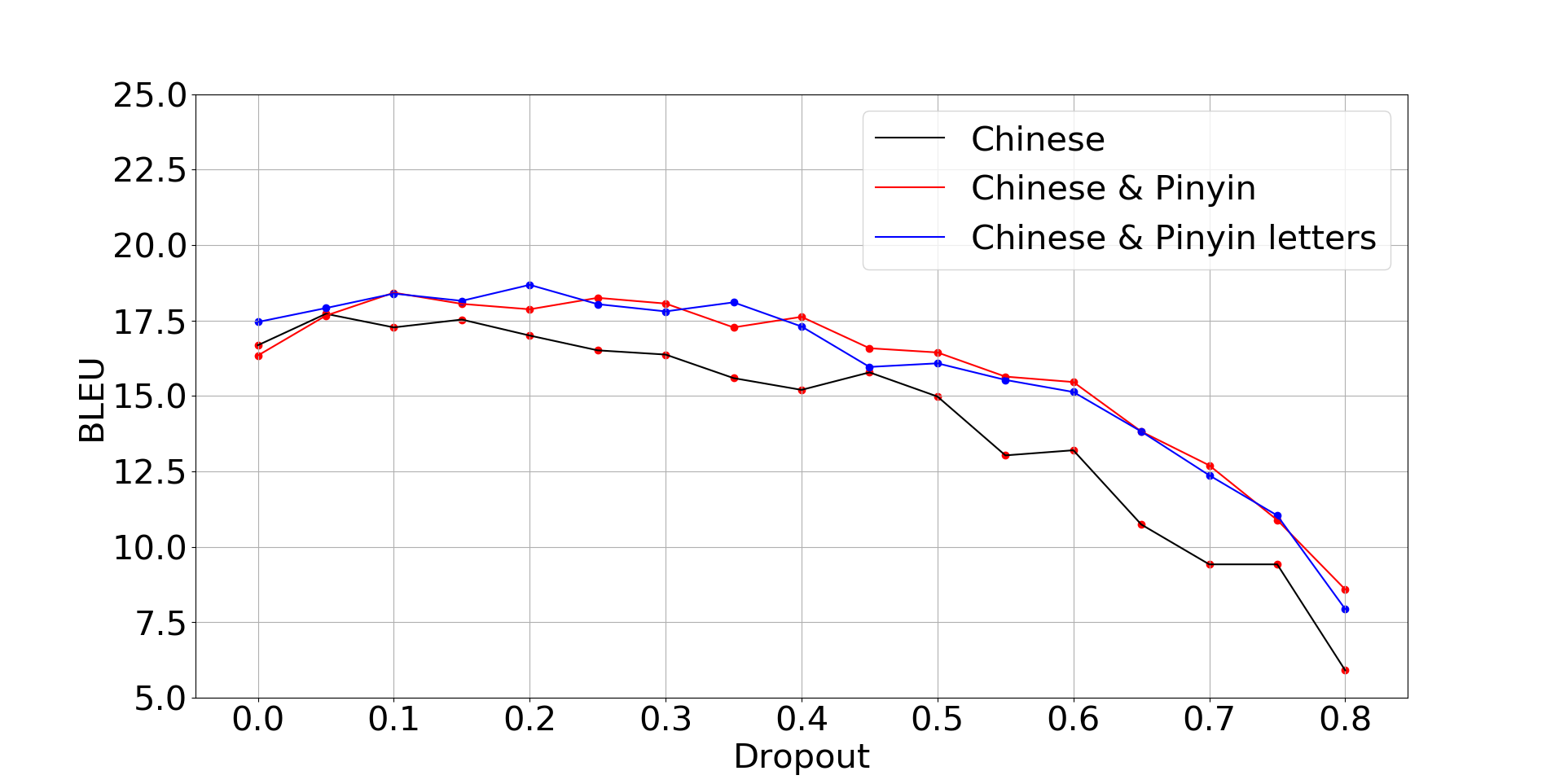}
    \caption{\small{Tuning for dropout. x-axis: the dropout value.}}
    \label{fig:dropout}
\end{figure}

We tune the dropout parameter for three experiments: Words, W+Pinyin, and W+Pinyin letters on ZH-EN. The drop out value is set by default to 0.2, and the beam-size to 12.  Figure \ref{fig:dropout} shows how translation accuracy changes by varying the dropout value. The highest BLEU score is at a dropout of 0.05 for the baseline, but between 0.2 and 0.3 for our approach. 
Higher optimial value of dropout means less nodes in the Neural Networks are needed to opt NMT quality. This implies that adding auxiliary inputs will reduce the model complexity. 
\begin{table}[t]
\begin{center}\scalebox{.7}{
\begin{tabular}{cc}
\begin{tabular}{c|cccccccc}
Coding &  FR-EN &  EN-FR & DE-EN & EN-DE  \\\hline
Words (W) &  48.7/26&  47.3/23  & 42.8/29& 48.5/26 \\\hline
Soundex & - &  66.5/30 & 37.8/27 & 50.3/25 \\
NYSIIS &  33.7/20 &  54.2/26 & 37.0/28 & 51.8/28 \\
Metaphone &  45.3/25 &  49.2/24  & 29.2/21 & 51.5/27\\\hline
W+Soundex &  - &  54.2/21  & 33.3/18 & 61.0/26\\
W+NYSIIS &  24.7/10 &  52.0/21 & 34.5/22 & 54.5/25 \\
W+Metaphone & \textbf{18.7}/8 & 65.3/26  & \textbf{33.0}/20 & 54.8/25\\
\end{tabular}
&
\begin{tabular}{c|c}
Coding &   ZH-EN \\\hline
Words (W) & 30.3/23 \\\hline
Pinyin &  45.8/32 \\
Wubi &  43.3/32 \\\hline
W+Pinyin & 49.2/26 \\
W+Wubi & 51.0/28 \\
\end{tabular}
\end{tabular}}\caption{\small{Training time in minutes/ number of epochs}} \label{tab:speed}
\end{center}
\end{table}

\subsection{Training Speed} 
Table \ref{tab:speed} shows the system training time (with BPE ${89k}$ operations and for ZH-EN ${18k}$). The total time (in minutes) is in the first column, and the number of epochs is in the second. 
The auxiliary information reduces the model complexity as described in Section~\ref{sec-complexity}. Therefore, the training becomes more efficient and needs a smaller number of epochs to converge. 
The total training time of our approaches is comparable to that of baselines, sometimes even less.

\begin{table}[t]
\begin{center}\scalebox{0.9}{
\begin{tabular}{c|cc}
Coding& ZH-EN \\\hline
Baseline: Words~\cite{Gehring17}& 17.00\\\hline
Wubi& 14.43\\
Pinyin & 15.57\\
Pinyin in letters & 12.51\\\hline
W+Wubi&18.11\\
W+Pinyin&17.87\\
W+Pinyin in letters & \textbf{18.68}\\\hline
K-Means characters& 14.57\\
random clustering words& 17.35\\
random clustering characters& 15.84\\\hline
W+K-Means words & 17.86\\
W+random clustering words& 18.47\\
W+random clustering characters& \textbf{18.49}\\
\end{tabular}}\caption{\small{Translation results  in BLEU[\%] for ZH-EN.
}}\label{tab-cnresult}
\end{center}
\end{table}

\subsection{Robustness}

\begin{table}[t]
\begin{center}\scalebox{0.7}{
\begin{tabular}{c|cccccc}
Source Data & \multicolumn{5}{c}{No. of Operations} \\ \cline{2-7} 
 & 0 & 1 & 2 & 3 & 4 & 5 \\ \hline
 Clean (C)  & 34.37 & 30.91 & 27.90 & 25.60 & 23.94  & 21.11 \\
 C+Soundex  & \textbf{35.88} & \textbf{32.31} & \textbf{28.72} & \textbf{26.52} & \textbf{24.51} & \textbf{22.27}     \\\hline
C+Noise (N) & 35.54 & 32.23 & 29.09 & 27.10 & 25.00 & 22.49     \\
C+N+Soundex & \textbf{35.80} & \textbf{32.28} & 28.83 & \textbf{27.14} & 24.81 & \textbf{22.76} \\
\end{tabular}}\caption{\small{Robustness in BLEU[\%]. Train: IWSLT'17 EN-FR adding noisy augmented data; Test: IWSLT'17 test by varying edit distance. }}\label{tab-robust2}
\end{center}
\end{table}

We test the system robustness with noise augmented data on IWSLT'17 EN-FR. 
In training, we randomly sample 20\% of words and substitute each of them with another word, which we randomly select according to its word2vec similarity to the substituted word. We concatenate the clean and noisy data and their phonetic encoded data. For each test sentence, we perform 0, 1, 2, 3, 4, or 5 times of the edit distance operation. We uniform randomly sample the word(s), the type of edit (from deletion, substitution, and insertion), and the substitution word. We trained models on below data: 1. Clean data (C); 2. Noisy data (N); 3. Noisy data concatenated with its Soundex encoding (N+S); 4. Combined clean and noisy data (C+N) as in~\cite{cheng2018towards}. Table~\ref{tab-robust2} (BPE 89k) shows that adding Soundex helps most test sets with different number of operations.

\begin{table}[t]
\begin{center}\scalebox{0.9}{
\begin{tabular}{c|ccccc}
Coding & MTNT'18 & MTNT'19 & WMT'15\\\hline
Baseline: Words&&&\\
~\cite{Gehring17} & 10.36 & 7.10 & 8.64 \\\hline
W+Soundex & 10.40 & 11.59 & 12.73 \\
W+NYSIIS & 10.58 & 10.67 & 13.39 \\
W+Metaphone & \textbf{10.98} & \textbf{12.65} & \textbf{14.53} \\
\end{tabular}}\caption{\small{Robustness in BLEU[\%] on an unknown test set oblivious during training. Train: IWSLT'17, EN-FR; Test: MTNT, WMT.}}\label{tab-robust}
\end{center}
\end{table}

Furthermore, we test the system robustness on a test set whose distribution is unknown during training (unlike the common robustness task). We aim to evaluate how a system behaves in a real-life scenario. We verified on English-French systems ($89k$) in Table~\ref{tab-result} trained on IWSLT'17. The tests are the out-of-domain News  WMT'15~\cite{Bojar2015findings} and informal language data MTNT test sets ~\cite{wmt2019robust} released the robustness shared task in WMT'19.
As in Table~\ref{tab-robust}, all of our approaches achieved higher accuracy than baselines. W+Metaphone outperforms all other systems and improves over the baseline by about 5 BLEU points.

\section{Related Work}

\cite {hayes1996phonetically,johnson2015sign} applied phonological rules or constrains to tasks such as word segmentation. 
Phonetics involves gradient and variable phenomena. Phonology has rules and patterns. In neural networks, we can directly learn from phonetic data and leave the network structure to discover hidden phonetic features opt NMT performance,  instead of optimizing towards phonological constraints. 
Discriminatively learning phonetic features has demonstrated success in various language technology applications. The work of~\cite{huang2004improving} used phonetic information to improve the named entity recognition task. ~\citet{bengio2014word} 
integrates speech information into word embedding and subword unit models, respectively. 
~\cite{du2017Pinyin} converted Chinese characters to subword units using Pinyin to reduce unknown words using a factor model. 
 Our work improves NMT overall rather than only translating unknown Chinese words.
We are the first to introduce the use of Soundex, NYSIIS, Metaphone, and Wubi in NMT. We successfully develop random clustering driven by our empirically verified hypothesis.

Leading research has investigated auxiliary information to NLP tasks, such as polysemous word embedding structures by  ~\citet{arora2016linear}, factored models by ~\citet{martnez2016factored} and~\citet{sennrich2016linguistic}, and feature compilation by  ~\citet{sennrich2016linguistic}. We applied both concatenation and multi-source encoding to combine phonetic inputs. 

Closely related, but independent to this work, is the approach of word segmentation or character-based NMT such as \cite{ling2015character,chung2016character}, which focuses on the decomposition of the translation unit. Smaller text granularity helps in unseen word challenges, while more extended translation units reduce input lengths. The use of Pinyin and logogram before BPE further chops down Chinese characters. Soundex, NYSIIS, and Metaphone have the effect of grouping letters. Both cases are different from the typical word segmentation task.
We take a different angle and view MT input as an information source encoded in various forms. We study the source sentence representations other than text such as phonetic encodings, which works surprisingly well when combined with word segmentation methods.  

\section{Conclusions}

We introduce phonetic and logogram encodings that hugely improve NMT translation quality (4 BLEU points) and robustness(5 BLEU points). When seeking the reasons for this improvement, we find and verify our hypothesis of diversity by phonetics, which leads to a new encoding algorithm, random clustering, that also significantly improves NMT. 

\bibliography{references}
\bibliographystyle{aaai}

\end{document}